\documentclass{article}

\usepackage[numbers]{natbib}
\usepackage[preprint]{neurips_2023}
\usepackage{graphicx} 
\usepackage{subcaption}

\usepackage[utf8]{inputenc} 
\usepackage[T1]{fontenc}    
\usepackage{url}            
\usepackage{booktabs}       
\usepackage{amsfonts}       
\usepackage{nicefrac}       
\usepackage{microtype}      
\usepackage{xcolor}         

\title{Reinforcement Learning for Dynamic Memory Allocation}

\author{%
  Arisrei Lim* \\
  University of Texas at Austin \\
  \texttt{arisrei@utexas.edu} \\
  \And
  Abhiram Maddukuri*\\
  University of Texas at Austin\\
  \texttt{abhicm@utexas.edu} \\
}

\usepackage[colorlinks=true, citecolor=blue]{hyperref}       

\begin{document}

\maketitle

\begin{center}
  Github: \url{https://github.com/arisrei/rl-malloc/}
  
  Video Summary: \url{https://youtu.be/m7EsBf1J5Pc}

  {\small * - equal contribution}
\end{center}

\begin{abstract}

In recent years, reinforcement learning (RL) has gained popularity and has been applied to a wide range of tasks. One such popular domain where RL has been effective is resource management problems in systems. We look to extend work on RL for resource management problems by considering the novel domain of dynamic memory allocation management. We consider dynamic memory allocation to be a suitable domain for RL since current algorithms like first-fit, best-fit, and worst-fit can fail to adapt to changing conditions and can lead to fragmentation and suboptimal efficiency. In this paper, we present a framework in which an RL agent continuously learns from interactions with the system to improve memory management tactics. We evaluate our approach through various experiments using high-level and low-level action spaces and examine different memory allocation patterns. Our results show that RL can successfully train agents that can match and surpass traditional allocation strategies, particularly in environments characterized by adversarial request patterns. We also explore the potential of history-aware policies that leverage previous allocation requests to enhance the allocator's ability to handle complex request patterns. Overall, we find that RL offers a promising avenue for developing more adaptive and efficient memory allocation strategies, potentially overcoming limitations of hardcoded allocation algorithms.

\end{abstract}

\section{Introduction}
Effective resource management is a pervading task in computer systems. It involves satisfying requests to use limited resources in a way that maximizes certain objectives (throughput, fairness, utilization, etc.). Examples of this problem include job scheduling on clusters, congestion control, CPU scheduling, and more. Such problems tend to be solved via hand-tuned, manually-designed, and empirically-tested heuristics. Thus, in many cases, management policies tend to be suboptimal and incur a high human workload to design and implement. This naturally leads to the idea of using machine learning as an alternative to heuristics.

Specifically, one can consider reinforcement learning (RL) \cite{sutton2018reinforcement} as a potential tool to learn algorithms. Recent success in RL for optimal control has led to its widespread in various contexts from video game AI to robotic manipulation. In fact, RL has already been used for effective resource management in computer systems. It has been used to find effective policies for tasks such as satisfying DRAM requests to maximize memory throughput \cite{ipek2008self}, job scheduling on servers with limited compute and memory \cite{mao2016resource}, and more.

While RL has been successfully applied to various resource management systems, one important domain it has not been applied to is dynamic memory allocation \cite{lea1996memory}. Dynamic memory allocation is a fundamental component of performance-critical systems; it allows programs to allocate memory as needed and control memory usage at a fine-grained level, ultimately giving programs the ability to limit their memory consumption. However, managing memory dynamically also introduces complexity in determining the optimal time and size of allocations to minimize fragmentation. Furthermore, traditional methods for dynamic memory allocation rely on algorithms such as first-fit, best-fit, or worst-fit, which can be suboptimal under varying load conditions. Additionally, these simple algorithms cannot adapt to varying system states, leading to decreased performance under a range of conditions. Alternatively, RL offers a promising approach to find a policy that can allocate blocks of memory efficiently. Through RL, a policy can be learned that makes optimal decisions on a per-state basis; thus, an allocator can make decisions with finer granularity, rather than applying the same hardcoded algorithm regardless of the state of the memory page. 


This paper explores the application of RL to dynamic memory allocation, proposing a framework where an RL agent continuously learns from system interactions to improve memory management. We first find that RL can be used to learn policies that directly output low-level address allocation actions. We also demonstrate that given adversarial request patterns for algorithms such as first-fit or best-fit, a policy can learn to avoid outputting suboptimal actions in those environments, and in some cases outperform the baselines. 
Our results lead us to the conclusion that RL is a promising avenue for learning effective memory allocation policies.

\section{Background and Related Work}

\subsection{Memory Allocation Problem}

In manually memory-managed languages, programs dynamically request memory via the malloc function and free that memory via the free function. If the region of memory returned by malloc is not carefully chosen, then memory can become fragmented; essentially, frequent allocations and deallocations can lead to memory fragmentation, where available memory is broken into small, unusable, non-contiguous blocks. This can lead to a severe underutilization of available memory when trying to allocate large contiguous memory blocks. Some of the most common solutions to the problem of memory allocation involve hard-coded logic to choose which region of memory to allocate to the user; however, these approaches do not adapt much depending on the state or history. Examples of such algorithms include the first fit, best fit, worst fit, and next fit algorithms. Indeed, the native version of malloc in some versions of Linux uses the best-fit algorithm \cite{lea1996memory}. We will use these algorithms as baselines for performance, and also as high-level actions for our policy to choose. 
 
While these “x-fit” hardcoded algorithms may have sufficient performance in the general case \cite{wilson1995dynamic}, there always exists a set of allocations that results in an inhibiting worst-case performance for the allocation algorithm; in other words, these hardcoded algorithms are inherently suboptimal. Applying RL to this problem enables us to gain two potential advantages over previous algorithms. Firstly, with enough training, a policy can learn to make optimal decisions on a per-state basis and thus can make more nuanced decisions. Secondly, given a history of states as input, our allocator can adapt temporal dependencies over allocation requests. More specifically, it is likely that individual requests in an allocation sequence are not independent. For example, consider a program with a loop that allocates a fixed memory amount in each iteration: Ideally, given a history of these requests, our allocator would learn to anticipate similar future requests and adapt accordingly.

\subsection{Reinforcement Learning for System Resource Allocation}

RL has become a common approach to solving problems involving the allocation of limited resources in a computer system \cite{tesauro2005online}. One such context involved using RL to learn a controller that satisfies DRAM requests while maximizing memory throughput \cite{ipek2008self}. This domain is especially relevant to our problem due to its many similarities. Firstly, the defined MDP for DRAM requests is stochastic due to outside factors (such as caching) influencing the received state. Additionally, the set of valid actions is dependent on the state. Despite these complications on the MDP, the agent was able to learn a policy that outperforms previous common algorithms, supporting the viability of using RL in our similar environment.

Another similar application involved using Deep RL for job scheduling on servers with limited compute and memory \cite{chen2017deep} \cite{mao2016resource}. Similar to us, both papers use a high-dimensional state space to represent the current state of allocated resources. Furthermore, both papers show the effectiveness of using neural networks and policy gradients in learning from this high-dimensional state space. One important difference to note is that all these works had access to the future request queue. However, in our problem of memory allocation, we do not have access to future memory requests.

While there exist many similar domains where RL is used, to the best of our knowledge this work is the first application of RL to dynamic memory allocation.

\subsection{Algorithms}
Our first set of algorithms apply to a state space of a single $page\_size$ bitmap vector representing the current allocation memory map. For this type of state, we first look to extract features (such as longest contiguous memory region, amount of free space, etc ), to use linear function approximation and Q-learning \cite{watkins1992q}. Additionally, following the success of \cite{chen2017deep} \cite{mao2016resource}, we will also evaluate directly passing in the bitmap and perform Deep RL using Deep-Q learning (DQN) and Proximal Policy Optimization (PPO) \cite{mnih2013playing} \cite{schulman2017proximal}. We chose DQN for high-level actions, since the state space is discrete and the number of actions is low. PPO is used for low-level actions, since the number of actions scales linearly with the state space. High vs low-level action spaces are expanded upon in \nameref{Problem Formulation}

Our next set of algorithms considers operating on a history of states. The simple way of doing this is to keep the last n memory allocation requests as part of the state, similar to \cite{mnih2013playing}. Our work will differ from this in that we don’t need to keep a history over the whole previous states, but only the previous memory allocation requests. We can then use this new history-aware state to train DQN policies. 

\section{Problem Formulation}
\label{Problem Formulation}

We cast the dynamic memory allocation problem as an MDP  with a tuple $(\mathcal{S}, \mathcal{A}, \mathcal{P}, \mathcal{R}, \gamma)$, where:
\begin{itemize}
  \item $\mathcal{S}$ is the set of states. For our environment, we consider states as a vector bitmap of size, $page\_size$ + 1. The first $page\_size$ elements are binary values of 1 and 0, where a value of 0 at index $i$ represents that memory address $i$ is unallocated and 1 represents allocated. The last value in the state corresponds to the requested allocation size. In addition to this fundamental state, we also experiment with appending the history of the last $n$ allocation amounts to see its effect on performance.
  
  \item $\mathcal{A}$ is the set of actions that the agent can take. We consider two separate action spaces. \begin{enumerate}
      \item \textbf{High-level actions}: Here, can take values between 0 and 2, where 0, 1, and 2 represent doing a first-fit, best-fit, and worst-fit allocation, respectively.
      \item \textbf{Low-level actions}: Here, actions are more granular and can take on values from 0 to $page\_size$-1. These numbers represent the exact index or address for the policy to allocate.
  \end{enumerate} 
  \item $\mathcal{P}$ is the state transition probability function: $\mathcal{P}(s' | s, a)$ represents the probability of transitioning to state $s'$ when taking action $a$ in state $s$. In our environment, transitions are stochastic because, in between subsequent states, an arbitrary number of free requests could have occurred causing the state to be changed in an unpredictable way.
  \item $\mathcal{R}$ is the reward function: $\mathcal{R}(s, a, s')$ represents the immediate reward received when transitioning from state $s$ to state $s'$ after taking action $a$. Our reward function simply returns 0.1 for every valid action taken. In the case of an invalid action being outputted (e.g. the low-level action network outputs an index which is already allocated), we give a reward of -10. If the largest free block in the page is less than the size of the allocation request, the episode ends since the request is unable to be fulfilled. This reward function encourages the policy to use the page's memory in the most efficient way possible by rewarding it for how long it is able to use the page.

  \item $\gamma$ is the discount factor, a scalar value between 0 and 1, which determines the agent's preference for immediate rewards over future rewards. In our environment we choose a discount factor of 1 since satisfying earlier allocation is no more important than satisfying later allocations.
\end{itemize}

The goal of RL is to find a policy $\pi : \mathcal{S} \rightarrow \mathcal{A}$ that maps states to actions, such that the cumulative expected reward is maximized:

\[
\pi^* = \arg\max_\pi \mathbb{E} \left[ \sum_{t = 0}^{\infty} \gamma^t \cdot \mathcal{R}(s_t, a_t, s_{t+1}) \right]
\]

where $s_t$ represents the state at time step $t$, $a_t$ is the action taken at time step $t$, and $\pi^*$ denotes the optimal policy.

\section{Experiments}

Through our experiments, we look to answer three questions
 \begin{enumerate}
     \item Can a policy that directly outputs low-level actions be learned at all?
     \item Given a sequence of adversarial request patterns for a certain first/best/worst-fit algorithm, can a policy learn not to pick those corresponding actions?
     \item Given a sequence of potentially adversarial or random request patterns, can a policy act optimally based on the request pattern type and ultimately outperform first/best/worst-fit baselines?
 \end{enumerate}

For the high-level action environments, unless otherwise stated, we use a history length of 10.

\begin{figure}
    \centering
    \includegraphics[width=0.5\linewidth]{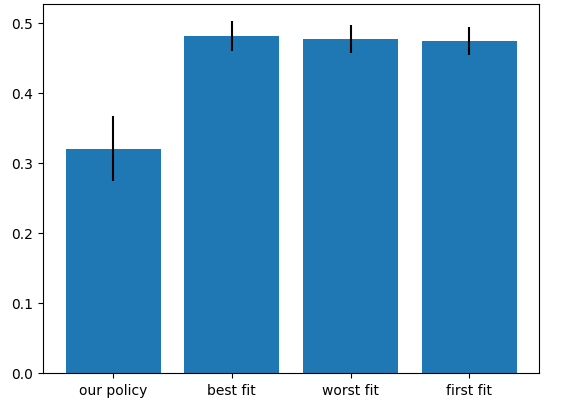}
      \caption{Average Return and 95\% Confidence Interval over 10000 Rollouts for Low-Level Action Network}
    \label{fig:lowlevel}

\end{figure}

\subsection{Experiment 1: Low-Level Action Network}
 For this experiment, we investigate the possibility of a policy to learn to directly output an index to allocate given an allocation request. We consider this to be more difficult than simply outputting high-level actions since the policy has to learn to associate allocation request sizes and the number of unallocated regions in the input in addition to finding optimal allocation schemes. For this reason, we initially start with $page\_size$ = 10; we use a random environment that frees with probability 0.4 and requests with probability 0.6; requests are randomly chosen from sizes 1-4. We parameterize our policy as a neural network and train using Stable Baselines 3 (SBB3) \cite{stable-baselines3} implementation of PPO. We use the the default hyperparameters given by SBB3 and train for 1000000 timesteps. 
 
 Examining our results in \ref{fig:lowlevel}, it is clear that the network eventually learns the association between the allocation request amount and valid allocation indices. Moreover, we see that the network learns a comparable policy to the other baselines in terms of return. However, we acknowledge the following caveats: first is that getting these results took 1 million timesteps of training; this may cause complications if training on larger page sizes requires an order magnitude more computation. Second, the policy still does not outperform the baselines. We hypothesize that this is due to 2 reasons. Firstly for a random request distribution, first/best/worst fit already achieves very strong results (to get better results we would need to train more). Second is that for 1 in every 500 episodes, our network may encounter a state which it does an invalid allocation for; since the network is deterministic it continues that action and receives a reward of around less than -40 at which point we end the episode. However, since the occurrence is so rare (about 1 in 500 episodes), we still find this result as encouraging evidence that a low-level action policy can be learned at larger scales.

 For the following experiments, we use a page size of 256 and only use policies with high-level action spaces.

 \begin{figure}
  \centering

  \begin{subfigure}[t]{0.32\textwidth}
    \centering
    \includegraphics[width=\textwidth]{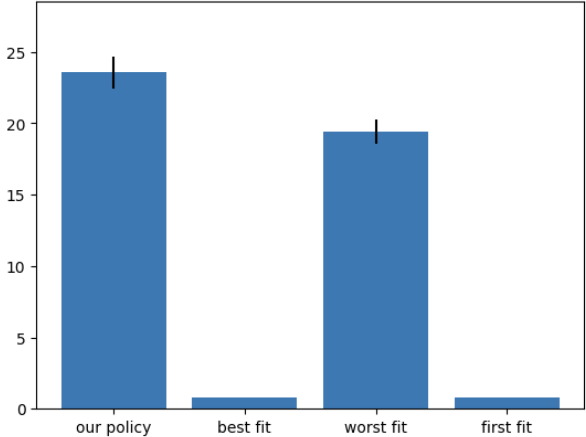}
    \caption{DQN + MLP on Wf-good}
    \label{fig:graph0}
  \end{subfigure}
  \hspace{0.01\textwidth}
  \begin{subfigure}[t]{0.325\textwidth}
    \centering
    \includegraphics[width=\textwidth]{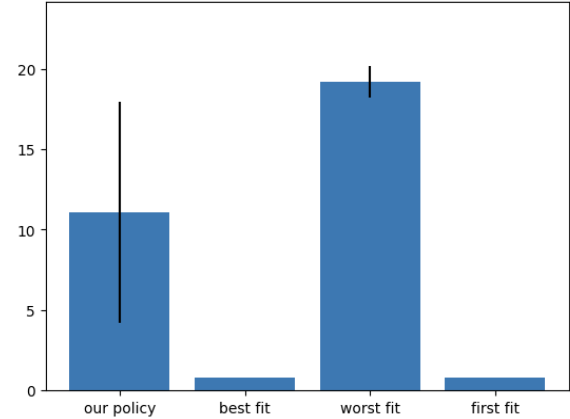}
        \caption{Linear Q-learning on Wf-good}
    \label{fig:linearwfgood}
  \end{subfigure}

  \begin{subfigure}[t]{0.32\textwidth}
    \centering
    \includegraphics[width=\textwidth]{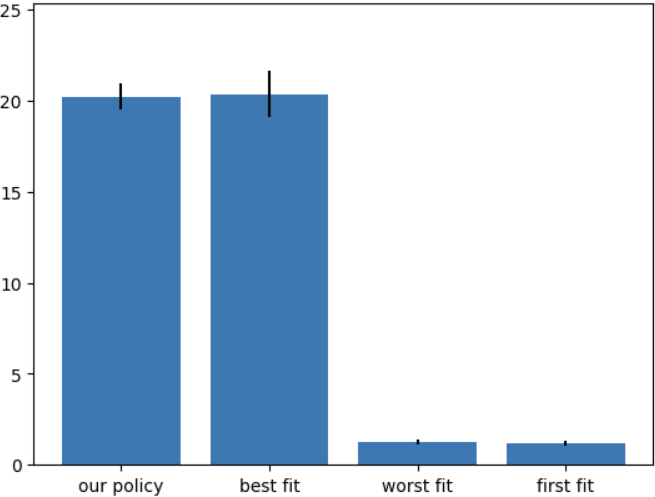}
    \caption{DQN + MLP on Bf-good}
    \label{fig:dqnmlpgfgood}
  \end{subfigure}
  \hspace{0.017\textwidth}
  \begin{subfigure}[t]{0.32\textwidth}
    \centering
    \includegraphics[width=\textwidth]{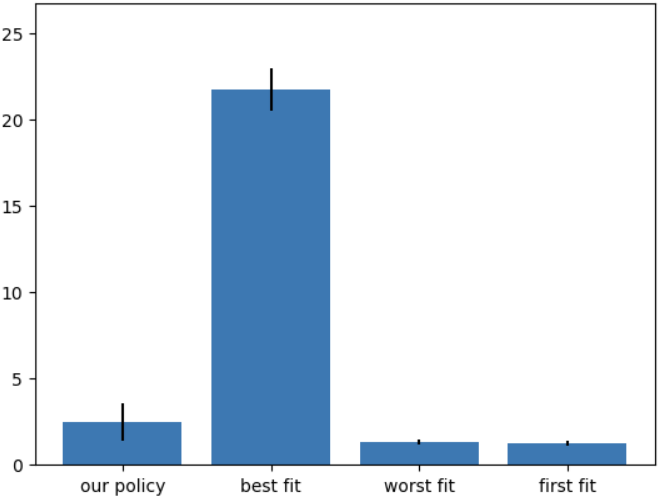}
    \caption{Linear Q-learning on Bf-good}
    \label{fig:graph3}
  \end{subfigure}

  \caption{Average Return and 95\% Confidence Interval over 5 training sessions, each with 100 Rollouts on Adversarial Allocation Requests}
  \label{fig:four_graphs}
\end{figure}

\subsection{Experiment 2: Adversarial Request Allocation}
For this experiment, we test an agent's performance on adversarial request patterns for the first/best/worst-fit algorithms. Specifically, we test performance against two different request generators: \textbf{bf-good} and \textbf{wf-good}. Bf-good draws from a distribution of request sequences that are unfavorable for first-fit and worst-fit allocation schemes and cause an episode end/allocation failure once the sequences are finished. However, a best-fit allocation scheme would be able to successfully satisfy all requests in the drawn sequence (Note once the drawn sequence is finished, our environment does random requests to not artificially end the episode). Wf-good is similar to bf-good except that the drawn sequences are unfavorable to first-fit and best-fit and favorable to worst-fit.

Our goal for these experiments is to evaluate whether or not a policy can learn to pick the favorable allocation corresponding to the adversarial request generators. We test this on 3 different policies, all of which use high-level actions. All policies were trained for 50000 timesteps. After 50000, policy performance did not seem to improve. 
\begin{enumerate}
    \item \textbf{Linear Q-Learning Policy}: For this policy, we train a linear action-value policy using Q-learning. We extract the following 9 features from our raw state: the average free block index and size, largest and smallest free block size, average allocated block index and size, total allocated block size, and largest and smallest allocated block size.
    \item \textbf{DQN + MLP Policy}: Here we train a fully-connected neural network using DQN. For hyperparameters, we use hidden layer sizes of [32, 32] and use a ReLU \cite{nair2010rectified} activation function between the hidden layers. We arrived at this architecture through trial and error, starting from the default hidden layer sizes of [64, 64], which performed suboptimally compared to our current architecture. We found the default learning rate of 0.0001 to be best, as higher learning rates don't seem to converge and lower ones don't converge fast enough. We also experimented with a 1D convolutional neural network rather than the previously described fully connected one. Because its performance was similar to the fully connected network, we have omitted its results from this paper. Interested parties can view its performance on the project GitHub page under results. Unless otherwise noted, we use the default hyperparameters from SBB3.

\end{enumerate}

Results for this experiment are shown at \ref{fig:four_graphs}. The results indicate that the neural network policies can learn to not choose first/worst/best fit when they are not appropriate for a respective allocation sequence. This confirms our initial hypothesis that, given a discernable pattern of requests, an agent can learn to pick optimal actions to maximize memory utilization. Moreover, looking at the results on the wf-good environment, we see evidence for the neural network outperforming the other baselines. This is only possible if the policy does not always pick the same "x-fit" action every time. Indeed, the distribution of picked actions has a majority on whichever action the request generator is designed for, but still picks other actions a significant amount of the time. This provides strong evidence that the memory allocation problem is not a simple bandit problem, as different high-level actions are better in different states even within the same environment created specifically to be good for one "x-fit" allocation scheme.

Looking at the linear policy, we notice that it performs well on bf-good but does not do well on wf-good. Due to the fact that neural networks performed well on both, we hypothesize that the linear policy's underperformance is likely due to a lack of informative features as opposed to using RL itself.

\begin{figure}
  \centering

  \begin{subfigure}[t]{0.4532\textwidth}
    \centering
    \includegraphics[width=\textwidth]{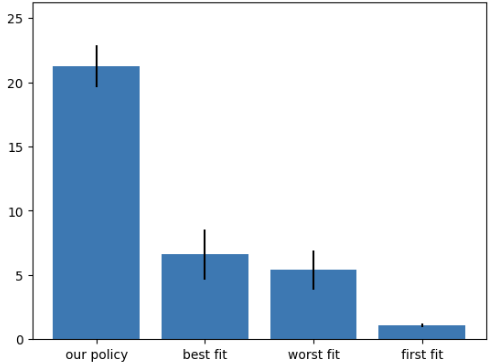}
    \caption{DQN + MLP Policy}
    \label{fig:graph1}
  \end{subfigure}
  \hspace{0.05\textwidth}  
  \begin{subfigure}[t]{0.4532\textwidth}
    \centering
    \includegraphics[width=\textwidth]{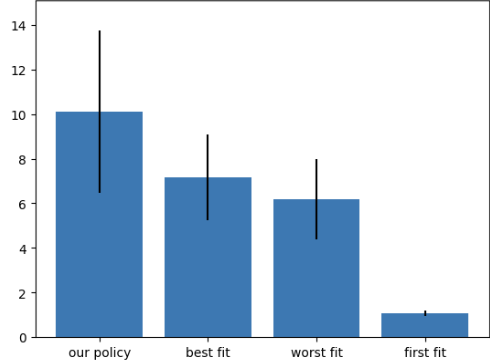}
    \caption{Linear Q-learning Policy}
    \label{fig:graph2}
  \end{subfigure}

  \caption{Average Return and 95\% Confidence Interval over 5 training sessions, each with 100 Rollouts on Mixed Allocation Requests}
  
  \label{fig:both_graphs}
\end{figure}

\subsection{Experiment 3: Mixed-Adversarial Allocation Requests}

For our final experiment, we test an agent's performance on a more complex series of "mixed" allocation requests on a page of size 256. Essentially, this environment gives allocation sequences that contain a random order of the adversarial allocation sequences and random allocation sequences. Such an allocation scheme is more reflective of real-world request patterns; thus, a policy that could adapt to the mixed-request environment would be more adaptive to the stochastic nature of real-world requests while also being robust to adversarial request patterns. To help the policy better adapt, we append the last 10 request sizes as a part of our state.

Results are shown in figure \ref{fig:both_graphs}. While the linear network shows potential to outperform baselines, we clearly see that the DQN + MLP policy does better than other baselines. More specifically, in the rollouts the DQN + MLP policy gets an average return of around 21, whereas the first/best/worst-fit baselines do not average more than 10. Considering we give a flat reward of 0.1 for each allocation timestep, this means the DQN + MLP policy is able to service requests for at least 100 more timesteps. While in practice an RL-trained policy would not attain this large of a performance boost since such adversarial patterns would not be as common/repeated, the results indicate a policy's ability to be robust to worst-case performance against adversarial requests while maintaining strong average-case performance with random allocations. 

\subsection{The role of history in our experiments}

While we test the effect of varying lengths of history in every low-level action experiment, we found that changing history length and keeping the type of function approximation architecture and type of request sequence constant did not make a significant difference in the average return. This gives some evidence that for our specific environments, a history of allocation requests does not provide useful information for the policy. We think this may be the case because we never explicitly define a pattern to be repeated in any of our environments. However, there are feasibly some allocation sequences in which a history of allocation requests would be useful for determining what type of action to take in the future. In addition, we used a constant history length for each combination. It is possible that a different history length we did not test would cause a change in policy performance. Future work to this end could consider using recurrent networks for function approximation as a way to remove the hyperparameter of history length.





\section{Limitations}

While our results illustrate the viability of using RL to learn robust and efficient policies, we acknowledge certain limitations in our results and approach and hope future work can be done to address them.

\begin{itemize}
    \item \textbf{Limited Scope}: The first limitation of note is a lack of full thoroughness in our experiments. Due to a lack of compute and limited time window, we did not get the chance to test across different seeds, do full hyperparameter searches, try a fuller suite of RL algorithms, or try various architectures. Additionally, with regards to experiment 1, we did not get a chance to train/test models for page sizes beyond 10 due to the fact that training that model already took 1000000 timesteps. Training for bigger page sizes would require even more compute. In addition, with more time and resources we could run more training sessions to decrease the sizes of our confidence intervals in experimental results.

    \item \textbf{Fidelity/Realism}: One point our work does not consider is the realizability of implementing our approach on a real system. We acknowledge that putting our approach into practice is a highly nontrivial task. Future work may include evaluations on the speed of our policy, as memory allocation is a very time-sensitive task. In addition, our work focuses on the performance of allocations on one page. We believe that with the correct adaptations, our algorithms can extend to environments with multiple pages.

    \item \textbf{Evaluation Data}: The last potential point of weakness is our use of simulation. While we believe our request simulator has aspects of realism and is informed by real-world patterns, we concede that it does not capture the full complexity of memory allocation. Ideally, we could use memory allocation patterns from common system programs to get a stronger evaluation. 
\end{itemize}

\section{Conclusion}

In this study, we cast and formulate dynamic memory allocation as a sequential MDP, demonstrating the capability of using RL to learn adaptive policies. We then create a simulated environment to evaluate RL agents on dynamic memory allocation. Our experiments show that with the correct set of hyperparameters and function approximations, our policies can match and outperform current first/best/worst fit algorithms when using various state and action spaces. Thus, we argue that RL presents as a potentially powerful tool in finding effective memory management policies. Despite our promising results, we acknowledge existing limitations in the thoroughness and scale of our benchmark, challenges in the practicality of system-level implementation, and overall drawbacks of using hand-engineered simulators. We leave these limitations to be addressed in future work.

\bibliographystyle{plain}
\bibliography{references}

\begin{thebibliography}{10}

\bibitem{chen2017deep}
Weijia Chen, Yuedong Xu, and Xiaofeng Wu.
\newblock Deep reinforcement learning for multi-resource multi-machine job
  scheduling.
\newblock {\em arXiv preprint arXiv:1711.07440}, 2017.

\bibitem{ipek2008self}
Engin Ipek, Onur Mutlu, Jos{\'e}~F Mart{\'\i}nez, and Rich Caruana.
\newblock Self-optimizing memory controllers: A reinforcement learning
  approach.
\newblock {\em ACM SIGARCH Computer Architecture News}, 36(3):39--50, 2008.

\bibitem{lea1996memory}
Doug Lea and Wolfram Gloger.
\newblock A memory allocator, 1996.

\bibitem{mao2016resource}
Hongzi Mao, Mohammad Alizadeh, Ishai Menache, and Srikanth Kandula.
\newblock Resource management with deep reinforcement learning.
\newblock In {\em Proceedings of the 15th ACM workshop on hot topics in
  networks}, pages 50--56, 2016.

\bibitem{mnih2013playing}
Volodymyr Mnih, Koray Kavukcuoglu, David Silver, Alex Graves, Ioannis
  Antonoglou, Daan Wierstra, and Martin Riedmiller.
\newblock Playing atari with deep reinforcement learning.
\newblock {\em arXiv preprint arXiv:1312.5602}, 2013.

\bibitem{nair2010rectified}
Vinod Nair and Geoffrey~E Hinton.
\newblock Rectified linear units improve restricted boltzmann machines.
\newblock In {\em Proceedings of the 27th international conference on machine
  learning (ICML-10)}, pages 807--814, 2010.

\bibitem{stable-baselines3}
Antonin Raffin, Ashley Hill, Adam Gleave, Anssi Kanervisto, Maximilian
  Ernestus, and Noah Dormann.
\newblock Stable-baselines3: Reliable reinforcement learning implementations.
\newblock {\em Journal of Machine Learning Research}, 22(268):1--8, 2021.

\bibitem{schulman2017proximal}
John Schulman, Filip Wolski, Prafulla Dhariwal, Alec Radford, and Oleg Klimov.
\newblock Proximal policy optimization algorithms.
\newblock {\em arXiv preprint arXiv:1707.06347}, 2017.

\bibitem{sutton2018reinforcement}
Richard~S Sutton and Andrew~G Barto.
\newblock {\em Reinforcement learning: An introduction}.
\newblock MIT press, 2018.

\bibitem{tesauro2005online}
Gerald Tesauro et~al.
\newblock Online resource allocation using decompositional reinforcement
  learning.
\newblock In {\em AAAI}, volume~5, pages 886--891, 2005.

\bibitem{watkins1992q}
Christopher~JCH Watkins and Peter Dayan.
\newblock Q-learning.
\newblock {\em Machine learning}, 8:279--292, 1992.

\bibitem{wilson1995dynamic}
Paul~R Wilson, Mark~S Johnstone, Michael Neely, and David Boles.
\newblock Dynamic storage allocation: A survey and critical review.
\newblock In {\em Memory Management: International Workshop IWMM 95 Kinross,
  UK, September 27--29, 1995 Proceedings}, pages 1--116. Springer, 1995.

\end{thebibliography}

\end{document}